\documentclass[final,5p,times]{elsarticle}

\usepackage[utf8]{inputenc}
\usepackage[T1]{fontenc}
\usepackage{url}
\usepackage{booktabs}
\usepackage{amsfonts}
\usepackage{nicefrac}
\usepackage{microtype}
\usepackage{soul}
\usepackage{color}
\usepackage{amssymb}
\usepackage{amsmath}
\usepackage{graphicx}
\usepackage{subfigure}
\usepackage{algorithm}
\usepackage{algorithmic}
\usepackage{multirow}
\usepackage{dblfloatfix}
\usepackage{balance}

\makeatletter
\def\ps@pprintTitle{%
 \let\@oddhead\@empty
 \let\@evenhead\@empty
 \def\@oddfoot{}%
 \let\@evenfoot\@oddfoot}
\makeatother

\journal{Engineering Applications of Artificial Intelligence}

\begin{document}

\begin{frontmatter}

\title{Random Error Sampling-based Recurrent Neural Network Architecture Optimization}

\author[uma,dlr]{Andr\'es Camero\corref{cor1}}
\ead{andrescamero@uma.es,andres.camerounzueta@dlr.de}
\cortext[cor1]{Correspondent author}

\author[uma,mit]{Jamal Toutouh}
\ead{toutouh@mit.edu}

\author[uma]{Enrique Alba}
\ead{ea@lcc.uma.es}

\address[uma]{Universidad de M\'{a}laga, ITIS Software, Espa\~{n}a}
\address[dlr]{GermanAerospace Center (DLR), Remote Sensing Technology Institute (IMF), Germany}
\address[mit]{Computer Science and Artificial Intelligence Laboratory, \\Massachusetts Institute of Technology, MA, USA}

\begin{abstract}
Recurrent neural networks are good at solving prediction problems. However, finding a network that suits a problem is quite hard because their performance is strongly affected by their architecture configuration.
Automatic architecture optimization methods help to find the most suitable design, but they are not extensively adopted because of their high computational cost.
In this work, we introduce the Random Error Sampling-based Neuroevolution (RESN), an evolutionary algorithm that uses the mean absolute error random sampling, a training-free approach to predict the expected performance of an artificial neural network, to optimize the architecture of a network.
We empirically validate our proposal on four prediction problems, and compare our technique to training-based architecture optimization techniques, neuroevolutionary approaches, and expert designed solutions. Our findings show that we can achieve state-of-the-art error performance and that we reduce by half the time needed to perform the optimization.
\end{abstract}


\end{frontmatter}

\noindent
\textbf{Copyright notice:} This article has been accepted for publication in the \textit{Engineering Applications of Artificial Intelligence} journal. Cite as: Camero, A., Toutouh, J., Alba, E. (2020). Random error sampling-based recurrent neural network architecture optimization. Engineering Applications of Artificial Intelligence, 96, 103946. \url{https://doi.org/10.1016/j.engappai.2020.103946}

\section{Introduction}\label{section:intro}

Deep learning (DL) and deep neural networks (DNN)~\cite{Lecun2015} are \emph{everywhere}, improving state-of-the-art results of problems from a wide range of topics, from natural language to image processing and pattern recognition~\cite{Litjens2017,Min2017}.

There are several types of DNNs, where each one is suited for solving a specific problem. Among these network types, Recurrent Neural Networks (RNNs) are especially good at solving sequential modeling and prediction, e.g., natural language, and speech recognition and modeling~\cite{Lecun2015}. Basically, RNNs are feedforward networks that include feedback connections between layers and neurons, and this recurrence allows them to capture long-term dependency in the input. In spite of their great performance, RNNs have a drawback: they are hard to train, because of the \emph{vanishing} and the \emph{exploding} gradient problems~\cite{bengio1994learning,Pascanu2013}. 

An alternative to mitigate these problems is to optimize the architecture of the network. By selecting an appropriate configuration of the parameters of the network (e.g., the activation functions, the number of hidden layers, the kernel size of a layer, etc.), it is tailored to the problem and by this means the performance is improved~\cite{Bergstra2011,Camero2018lion,Jozefowicz2015}. 

DNN architecture optimization methods can be grouped into two main groups: the manual exploration-based approaches (usually) lead by expert knowledge, and the automatic search-based methods, e.g., grid, neuroevolutionary or random search~\cite{Ojha2017}. 

The number of alternatives or parameters to configure a DNN is extremely large. Thus the architecture optimization has to deal with a high-dimensional search space. Despite this size, most methods (manual and automatic) are based on \textit{trial-and-error}, meaning that each architecture is trained (e.g., using a gradient-based algorithm) and tested to evaluate its numerical accuracy. Thus, the high-dimensional search space and the high cost of the evaluation limit the interest in this methodology~\cite{Ojha2017}. 

Some authors have explored different approaches to speed up the evaluation of DNN architectures to improve the efficiency of automatic architecture optimization algorithms~\cite{Camero2018lowcost,Domhan2015}. Among them, the \emph{Mean absolute error random sampling} (MRS)~\cite{Camero2018lowcost,camero2018comparing} poses a different way of dealing with the problem. The main idea behind this method, inspired by the linear time-invariant theory, is to infer the numerical accuracy of a given network without actually training it. Given an input, several sets of random weights are generated and analyzed measuring the mean absolute error (MAE). Then, the probability of finding a set of weights whose MAE is below a predefined threshold is estimated. 

Therefore, can we combine the MRS (to speed-up the search) and a gradient-based technique (to improve the performance) into a \emph{hybrid} technique to optimize the architecture of an RNN? More specifically, we pose the following research questions:

\begin{description}
    \item[\textbf{RQ1}]  Can a \emph{hybrid} (MRS and gradient-based) architecture optimization technique get the same error performance of a solely gradient-based one?
    \item[\textbf{RQ2}] Can a \emph{hybrid} architecture optimization technique get the same error performance of a non-gradient-based approach (.e.g., neuroevolutionary algorithm)?
    \item[\textbf{RQ3}] Can we reduce the architecture optimization time by using this \emph{hybrid} approach?
    \item[\textbf{RQ4}] Can we improve the performance of an expert designed architecture/solution?
\end{description}

To answer this questions, we propose a novel technique: the Random Error Sampling-based Neuroevolution (RESN), an evolutionary algorithm (EA) that navigates through the architecture space and guides its search by using the MRS, avoiding the high-cost of training each candidate solution. Then, once the algorithm has computed a ``final'' solution, RESN will train it using a gradient descent-based method.

Therefore, the main contribution of this work is to propose a \emph{hybrid} approach to optimize the architecture of a RNN, i.e., an evolutionary algorithm that relies on the MRS as a fitness measure to select an RNN architecture, and uses a gradient-based technique to train the \emph{final} architecture.

The remainder of this paper is organized as follows: the next section briefly reviews related works. Section~\ref{section:optimization} introduces RESN. Section~\ref{section:setup} presents the experimental design. Section~\ref{section:results-optimization} presents the results. And finally, Section~\ref{section:conclusions} outlines the conclusions and proposes future work.

\section{Related Work}\label{section:related}

This section outlines the related work. First, we introduce RNN architecture optimization and evaluation works. Second, we briefly review the MRS. Finally, we review the state-of-the-art of metaheuristics applied to DL optimization.

\subsection{RNN Architecture Optimization}

An RNN is an artificial neural network that adds recurrent (or feedback) edges that may form cycles and self connections~\cite{Lipton2015}. Due to this recurrency, most gradient-based optimization procedures fail to train an RNN. The main issue with gradient-based approaches is that they keep a vector of activations, which makes RNNs extremely deep and aggravates the exploding and the vanishing gradient problems~\cite{bengio1994learning,Pascanu2013,Kolen2001}.    

To tackle these gradient-related problems, Hochhreiter and Schmidhuber proposed an effective solution: the Long Short-Term Memory (LSTM) cell. The LSTM is a special \emph{neuron design} that contains units called memory blocks in the recurrent hidden layer~\cite{Hochreiter1997}. Despite LSTM effectively mitigates gradient problems (i.e., they are easier to train than standard RNNs), not only the network architecture affects the learning process but also the weight initialization~\cite{Ramos2017} and all specific parameters of the optimization algorithm~\cite{haykin2009neural}.

Therefore, to cope with the learning process as a whole, some authors have proposed to perform an architecture optimization~\cite{Bergstra2011,Camero2018lion,Jozefowicz2015}. Specifically, they propose to look for a specific architecture (the number of layers, the number of hidden unit per layer, etc.) and a set of parameters to train the network that improve the performance of the \emph{optimized} network given a data set. In other words, instead of using a general configuration, the idea is to tailor the architecture to the problem.

When dealing with an architecture configuration, an expert can discard a configuration based on his expertise, i.e., without the need of evaluating it. However, intelligent automatic architecture optimization procedures search more efficiently through a high-dimensional search space.

Even though intelligent methods are more competitive than experts, they are not generally adopted because they are computationally intensive~\cite{Ojha2017}. They require to fit a model and to evaluate its performance on validation data (i.e., they are data-driven), which can be a demanding process (time and computational resources)~\cite{Bergstra2011,Albelwi2017,Smithson2016}.

Hence, few methods have been proposed to address this issue by speeding up the evaluation of the proposed architecture. For example, Domhan et al.~\cite{Domhan2015} proposed to predict the performance of a network based on the learning curve, reducing the search time up to 50\%. 

More recently, Camero et al.~\cite{Camero2018lowcost} presented the MRS, a novel low-cost method to compare the performance of RNN architectures without training them. Particularly, the MRS evaluates a candidate architecture by generating a set of random weights and evaluating its performance. 

In line with the latter approach, we propose to use the MRS to guide an EA to search for the most suitable architecture.

\subsection{Mean Absolute Error Random Sampling}

At a glance, the MRS predicts how \emph{easy} would be to train a network (i.e., the performance) by taking a user-defined number of samples of the output (on a given input) of a specific architecture. Each sample is taken using a (new) random normally distributed set of weights, and calculating the MAE. Then, a truncated normal distribution is fitted to the MAE values sampled, and a probability $p_t$ of finding a set of weights whose error is below a user-defined threshold is estimated. Then, the probability $p_t$ is used as a predictor of the performance (error) of the analyzed architecture.

Algorithm~\ref{algorithm:sampling} (taken from~\cite{Camero2018lowcost}) presents the pseudo-code of the MRS. Given an architecture (\textit{ARQ}), a number of time steps (or look back, \textit{LB}), and a user-defined input (\textit{data}), the algorithm initializes the architecture (\textbf{InitializeRNN}). Then, it takes \textit{MAX\_SAMPLES} samples of the MAE, i.e., for each sample a set of normally distributed weights is generated (function \textbf{GenerateNormalWeights}), with mean equal to zero, and standard deviation equal to one. The RNN is updated with the new set of weights (\textbf{UpdateWeights}), and the MAE is computed for the data using the updated RNN (\textbf{MAE}). Once the sampling is done, a truncated normal distribution is fitted to the MAE values sampled (\textbf{FitTruncatedNormal}), and finally, the probability $p_t$ is estimated for a defined \textit{THRESHOLD} (\textbf{PTruncatedNormal}).

The details and the design considerations, including the tuning of the parameters, are thoroughly discussed in the original proposal~\cite{Camero2018lowcost}.

\begin{algorithm}[!h]
	\caption{MRS pseudo-code~\cite{Camero2018lowcost}}
	\label{algorithm:sampling}
	\begin{algorithmic}[1]
	    \STATE {\textbf{Given}: an architecture (\textit{ARQ}), a number of time steps or look back (\textit{LB}), a user-defined time series (\textit{data}), a number of samples (\textit{MAX\_SAMPLES}), and a \textit{THRESHOLD}.}
		
        \STATE {$\text{rnn} \gets \operatorname{InitializeRNN}(\text{ARQ}, \text{LB})$}
        \STATE {$\text{mae} \gets \emptyset$}
        \WHILE {$\text{sample} \leq \text{MAX\_SAMPLES}$} 
            \STATE {$\text{weights}  \gets  \operatorname{GenerateNormalWeights}(\mu = 0, \sigma = 1)$}
            \STATE {$\operatorname{UpdateWeights}(\text{rnn}, \text{weights})$}
            \STATE {$\text{mae}[\text{sample}] \gets \operatorname{MAE}(\text{rnn}, \text{data})$}	
            \STATE {$\text{sample}++$}
		\ENDWHILE
        \STATE {$\text{mean}, \text{sd} \gets \operatorname{FitTruncatedNormal}(\text{mae})$}
        \STATE {$p_t \gets \operatorname{PTruncatedNormal}(\text{mean}, \text{sd}, \text{THRESHOLD})$ }
	\end{algorithmic}
\end{algorithm}

\subsection{Deep Learning and Metaheuristics}

Metaheuristics are well-known optimization algorithms to address complex, non-linear, and non-differentiable problems~\cite{Ojha2017,back1996evolutionary}. 
They efficiently combine exploration and exploitation strategies to provide \emph{good} solutions requiring bounded computational resources. 

Optimization in DL may be viewed from different perspectives: training as optimization of the DNN weights, hyperparameter selection, network topologies, learning environment, etc. 
These different points of view are adopted to improve the DNNs generalization capabilities. 

Gradient-descent based methods, such as back-propagation, are widely used to train DNNs. However, these methods need several manual tuning schemes to make their parameters optimal and it is difficult to parallelize them taking advantage of graphics processing units (GPUs). 
Thus, several authors have explored DNNs training by using metaheuristics, an idea explored long before DNN rise~\cite{alba1993full,alba2006metaheuristic}. Different authors combined convolutional neural networks with metaheuristics to improve their accuracy and performance by optimizing the layers weights and threshold. Following this idea, Zhining and Yunming~\cite{zhining2015genetic} used a genetic algorithm (GA); Rosa et al.~\cite{10.1007/978-3-319-25751-8_82} applied harmony search (HS); Rere et al.~\cite{RERE2015137} analyzed simulated annealing (SA), and later the same authors evaluated SA, differential evolution (DE), and HS~\cite{rere2016metaheuristic}. 

Some authors also explored coupling training based on stochastic gradient descent (SGD) with metaheuristics. This approach has provided promising results in training generative adversarial networks (GANs)
GANs combine a generative network (generator) and a discriminative network (discriminator) that apply adversarial learning to be trained.
Evolutionary GAN (E-GAN) applies ES to evolve a population of networks (generators) that are mutated by applying SGD according to different loss functions~\cite{wang2019evolutionary}. The generators are evaluated by a single discriminator that returns a fitness value for each generator. 
Lipizzaner~\cite{schmiedlechner2018lipizzaner} and Mustangs~\cite{toutouh2019} are competitive coevolutionary algorithms that evolve two populations, one of generators and one of discriminators, to improve diversity during the training. They also apply SGD-based mutation to generate the offspring.

GA has been applied to evolve increasingly complex neural network topologies and the weights simultaneously, in the NeuroEvolution of Augmenting Topologies (NEAT) method~\cite{stanley2002evolving,larochelle2009exploring}. However, NEAT has some limitations when it comes to evolving DNNs and RNNs~\cite{miikkulainen2019evolving}.

Focusing on RNNs, NEAT-LSTM~\cite{rawal2016evolving} and CoDeepNeat~\cite{Liang:2018:EAS:3205455.3205489} extend NEAT to mitigate its limitations when evolving the topologies and weights of the network. Besides, particle swarm optimization (PSO) has been analyzed to train RNNs instead of SGD~\cite{IBRAHIM2018216}, providing comparable results. ElSaid et al.~\cite{ElSaid2018} proposed the use of ant colony optimization (ACO) to improve LSTM RNNs by refining their cellular structure. Ororbia et al.~\cite{ororbia2019investigating} introduced the Evolutionary eXploration of Augmenting Memory Models (EXAMM), which is capable of evolving RNNs using a wide variety of memory structures. ElSaid et al.~\cite{elsaid2019evolving} introduced the Evolutionary eXploration of Augmenting LSTM Topologies (EXALT), a techniques for evolving RNNs, including epigenetic weight initialization and node-level mutation operations.  Camero et al.~\cite{Camero2018lion} applied GA to search for the most efficient ones to improve the accuracy and the performance regarding the most commonly used RNNs configurations. In this case, the authors train the network using SGD to evaluate the performance of the configurations. 
As a conclusion of this literature review, it seems that the main difference between our approach and all the mentioned metaheuristics is that we propose to use the MRS instead of training each network/configuration. Thus, we expect to reduce the computational cost of the evaluation process, allowing the optimization algorithm to perform a larger number of iterations.

\section{Random Error Sampling-based Neuroevolution}\label{section:optimization}

In this section, we introduce RESN, our proposal for RNN architecture optimization based on the MRS. First, (\emph{i}) we state the architecture optimization problem and then, (\emph{ii}) we present an evolutionary algorithm to perform such optimization.

\subsection{Architecture Optimization}
\label{sec:arq-optimization}

The optimization of the architecture of an artificial neural network consists of searching for an appropriate network structure (i.e., the architecture) and a set of weights~\cite{haykin2009neural}. However, in spite of this definition, it is rather common to arbitrarily define the architecture and then applied a learning rule (e.g., SGD) to optimize the set of weights~\cite{Ojha2017}. Thus, we might say that the network is partially optimized or, in other words, we are not fully leveraging the computational model.

Usually, the RNN architecture optimization is stated as a minimization problem~\cite{Ojha2017}. For example, we may define this problem as looking for an RNN architecture that minimizes the mean absolute error (MAE) of the predicted output ($z_i$) against the real one ($y_i$), subject to a minimum/maximum architecture (ARQ) definition (i.e., the number of hidden layers, the number of neurons per each layer, and the connecting edges), and to a minimum/maximum look back (LB). The training of the candidate solution is usually implied in this definition. Therefore, due to the intensive computations of the training, this optimization tends to be time demanding. Therefore, we propose to reformulate the optimization problem using the MRS.

We propose to optimize the architecture of an RNN by maximizing $p_t$, i.e., given an input \emph{X} and an output \emph{Y}, we propose to look for an RNN architecture that maximizes the estimated probability of finding a set of weights whose error is below a user-defined threshold (Algorithm~\ref{algorithm:sampling}). Equation~\ref{eq:maximization} presents the referred problem.

\begin{align}\label{eq:maximization}
\text{maximize }  & \text{Heuristic} = p_{t}(X, Y)\\
\text{subject to } 
    & \text{min\_ARQ} \le ARQ \le \text{max\_ARQ}  \nonumber \\
    & \text{min\_LB} \le LB \le \text{max\_LB} \nonumber 
\end{align}

\subsection{Evolutionary Approach}

To solve the RNN architecture optimization problem stated in Equation~\ref{eq:maximization}, we designed a \emph{deep neuroevolutionary} algorithm based on the ($\mu+\lambda$) EA~\cite{back1996evolutionary}. Algorithm~\ref{algorithm:optimizer} presents a high-level view of our proposal.

\begin{algorithm}[!h]
	\caption{Random Error Sampling-based Neuroevolution}
	\label{algorithm:optimizer}
	\begin{algorithmic}[1]
	    \STATE {population $\gets$ Initialize(\textit{population\_size})}
        \STATE {Evaluate(population)}
        \STATE {evaluations $\gets$ \textit{population\_size}}
        \WHILE {evaluations $\leq$ \textit{max\_evaluations}} 
            \STATE {offspring $\gets$ BinaryTournament(population, \textit{offspring\_size})}
            \STATE {offspring $\gets$ CellMutation(offspring, \textit{cell\_mut\_p}, \textit{max\_step})}
            \STATE {offspring $\gets$ LayerMutation(offspring, \textit{layer\_mut\_p})}
            \STATE {Evaluate(offspring)}
            \STATE {population $\gets$ Best(population + offspring, \textit{population\_size})}
            \STATE {evaluations $\gets$ evaluations + \textit{offspring\_size}}
            \STATE {SelfAdapting(\textit{layer\_mut\_p}, \textit{max\_step}, \textit{cell\_mut\_p})}
		\ENDWHILE
        \STATE {solution $\gets$ Best(population, 1)}
        \STATE {rnn\_trained $\gets$ Train(solution, \textit{epochs})}
        \STATE {\textbf{return} rnn\_trained}
	\end{algorithmic}
\end{algorithm}

\sloppy
Here, a solution represents an RNN architecture (i.e., ARQ in Equation~\ref{eq:maximization}), and it is encoded as a variable length integer vector, \mbox{$sol = < s_0, s_1, ..., s_H >$}, where $s_0$ is the LB \mbox{$s_0 \in [\text{min\_LB},\text{max\_LB}]$}, and $s_i$ ($i \in [1,\text{H}]$) corresponds to the number of LSTM cells in the $i$-th hidden layer. Thus, \mbox{$s_i \in [\text{min\_NPL},\text{max\_NPL}]$} and \mbox{$H \in [\text{min\_HL},\text{max\_HL}]$}. Given this definition, the number of hidden layers is implicitly derived from the length of the solution. Then, the \textbf{population} is defined as a set of \emph{population\_size} solutions.

First, the \textbf{Initialize} function randomly creates a set of solutions (using a uniform distribution). Next, the \textbf{Evaluate} function computes $p\_t$ (Algorithm~\ref{algorithm:sampling}) for each solution. Then, the population is evolved until the termination criteria is met (i.e., the number of evaluations is greater than \emph{max\_evaluations}). 

The evolutionary process is divided into selection, mutation, evaluation, replacement, and self-adjustment (Algorithm~\ref{algorithm:optimizer}). First, (line~5) an \textbf{offspring} (of \emph{offspring\_size} solutions) is selected using a binary tournament from the actual population. Then, each solution in the offspring is mutated by a two step process. In the first step of the mutation (line~6, \textbf{CellMutation}), for every $s_j$ (\mbox{$j \in [0,H]$}), with a probability \emph{cell\_mut\_p}, a value in the range \mbox{$[\text{min\_step},\text{max\_step}]$} (excluding zero) is added. Then, in the second step of the mutation (line~7, \textbf{LayerMutation}), independently, with a probability \emph{layer\_mut\_p}, the layer $s_i$ ($i \in [1,\text{H}]$) is cloned or removed (with the same probability), i.e., one layer is added or subtracted to the solution.

Once the mutation is done, the offspring is evaluated (\textbf{Evaluate}) using the MRS, and after, the best solutions from the population and the offspring are selected by the \textbf{Best} function, i.e., the population and the offspring are gathered together, sorted, and finally, the solutions that have a higher $p_t$ give place to the new population (of \emph{population\_size} solutions). 

The number of evaluations is increased by \emph{offspring\_size}. And finally, a \textbf{SelfAdapting} process takes place. In this process, if the new population is improving on average (i.e., the average $p_t$ of the new population is greater than the former average), then, the \emph{cell\_mut\_p}, \emph{max\_step}, and \emph{layer\_mut\_p} parameters are multiplied by 1.5. Otherwise, these parameters are divided by 4. These numbers/values are taken from~\cite{Doerr17tutorial}.

After the evolutionary process ends, the best solution (i.e., the solution with the greatest $p_t$) of the population is selected (line~13), and trained using a user-defined method. Without loss of generality, we defined to use Adam~\cite{kingma2014adam} optimizer to the train the final solution for a predefined number of \emph{epochs}. 

\begin{figure}[!h]
    \centering
    \includegraphics[width=0.8\columnwidth]{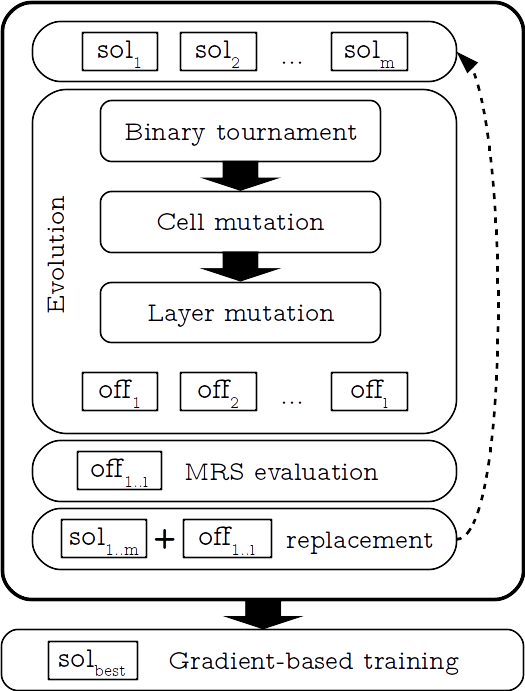}
    \caption{The global scheme of RESN}
    \label{fig:resn}
\end{figure}

Finally, the algorithm returns an RNN that is optimized (structure and weights) to the given problem. Figure~\ref{fig:resn} depicts a high-level view of RESN.

It is quite interesting to notice that the \textbf{Evaluate} function may be changed seamlessly by any other fitness function, e.g., the MAE after training the network for a user-defined number of times. Accordingly, the \textbf{Best} function has to be modified to maximize or minimize the new objective function (fitness).

\section{Experimental Setup}\label{section:setup}

This section introduces the study performed. First, we present and justify the data sets. Second, we introduce the experiments carried out to test our approach.

\subsection{Data Sets}

To test our proposal and answer the research questions we selected four problems: the sine wave, the waste generation prediction problem~\cite{Ferrer2018Bio}, the coal-fired power plant flame intensity prediction problem~\cite{ororbia2019investigating}, and the load forecast problem~\cite{chen2004load}.

The \emph{sine wave} problem consists of predicting the following value of the function, given the historical data. A sine wave is a periodic oscillation curve, that may be expressed as a function of time (\emph{t}), where $A$ is the peak amplitude, $f$ is the frequency, and $\phi$ is the phase (Equation~\ref{equation:sine}). Particularly, we used the sine wave described by $A=1$, $f=1$, and $\phi=0$, in the range $t \in [0,100]$~seconds (s), with ten samples per second. In spite of its simplicity, this problem is very useful because any periodic waveform can be approximated by adding sine waves and it is extensively used in the literature~\cite{Camero2018lowcost,camero2018comparing,bracewell1986fourier}.

\begin{align}\label{equation:sine}
y(t) = A \cdot sin(2 \pi \cdot f \cdot t + \phi)
\end{align}

The \emph{waste generation prediction problem}, introduced by Ferrer and Alba~\cite{Ferrer2018}, consists of predicting
the filling level of 217 recycling bins located in the metropolitan area of a city in Spain. Each filling level is recorded daily for one year. Therefore, given the historical data (i.e., 217 input values per day), the problem is to predict the next day (i.e., the filling level of all bins). 

It is important to notice that this problem was originally proposed to predict the filling level of each container individually using \emph{Gaussian processes}, \emph{linear regression}, and \emph{SMReg}~\cite{Ferrer2018}. A later work (we will refer to this results as \emph{Short training}) outperformed those results by predicting all containers at once using one single RNN~\cite{Camero2018Waste}.

The third validation case is the \emph{coal-fired power plant flame intensity prediction problem}. It was introduced by Ororbia et al.~\cite{ororbia2019investigating}, and consists of ten days of values recorded every minute for 12 burners. The original data set has been pre-normalized to the range [0,1]. The problem is to predict the main flame intensity given the historical data.

Finally, the \emph{load forecast problem}~\cite{chen2004load}, originally introduced by EUNITE network in the 2001 competition called ``Electricity Load Forecast using Intelligent Adaptive Technologies''\footnote{\url{http://www.eunite.org/}}, consists of a mid-term load forecasting challenge. Particularly, the data set includes the electricity load demand of the Eastern Slovakian Electricity Corporation every half hour, from January 1, 1997, to January 31, 1999. As well as the temperature (daily mean), and the working calendar for the referred period. Then, using the data from 1997 to 1998, the challenge is to predict the maximum daily load of the January 1999 days.

\subsection{Experiments}

To answer the research questions we propose to implement RESN and benchmark it against state-of-the-art techniques to optimize the architecture of an RNN. Particularly, we propose the following four experiments.

\subsubsection{E1: RESN vs. Gradient-based Architecture Optimization}

To answer the question of whether a \emph{hybrid} (MRS and gradient-based) architecture optimization technique can get the same error performance of a solely gradient-based one, we propose three tests. In all cases, based on the MRS original proposal~\cite{Camero2018lowcost}, we set the MRS parameters according to Table~\ref{table:sampling-params}. It is worth noticing that the parameters set in the referred table may influence the results. Nonetheless, we decided to rely on the values set in the original paper, as tuning them is not in the scope of this study.

\begin{table}[!h]
    \caption{MRS parameters}
	\centering 
    \begin{tabular}{ lr }
    \hline
    Parameter & Value \\
    \hline
    MAX\_SAMPLES    & 100 \\
    THRESHOLD   & 0.01 \\
    \hline
	\end{tabular} \\    
    \label{table:sampling-params}
\end{table}

First, for the sine wave, we propose to (\emph{E1.i}) run Algorithm~\ref{algorithm:optimizer} but using the \emph{early} training results to evaluate the solutions, i.e., we propose to train each candidate architecture using Adam for a short time (one epoch), use the trained architecture to predict on the test data set, compute the MAE, and use that value as the heuristic of the optimization algorithm (Algorithm~\ref{algorithm:optimizer}, line 8). For this test, we set the parameters according to Table~\ref{table:test-i-params}. It is very important to notice that during the optimization \emph{cell\_mut\_p, max\_step, and layer\_mut\_p} values are self-adapted (Algorithm~\ref{algorithm:optimizer}, line~11). Thus, their initial values are not critical~\cite{Doerr17tutorial}.

\begin{table}[!h]
    \caption{Parameters of E1.i: RESN vs. Gradient-based architecture optimization}
	\centering 
    \begin{tabular}{ lrc|clr }
    \hline
    Param & Value &
        && Param & Value \\
    \hline
    cell\_mut\_p & 0.2 &
        && min\_LB & 2 \\
    epochs & 100 &
        && max\_LB & 30 \\
    max\_step & 5 &
        && min\_NPL & 1 \\
    layer\_mut\_p & 0.2 &
        && max\_NPL & 100 \\
    population\_size & 10 &
        && min\_HL & 1 \\
    offspring\_size & 10 &
        && max\_HL & 3 \\
    max\_eval & 100 &
        && & \\
    \hline
	\end{tabular} \\    
    \label{table:test-i-params}
\end{table}

Second, for the waste prediction problem, we propose to (\emph{E1.ii}) benchmark our results against \emph{Short training}~\cite{Camero2018Waste}. Particularly, the authors proposed to optimize an RNN to the problem using an ES-based algorithm. More specifically, they trained each candidate solution using \emph{Adam}~\cite{kingma2014adam} for a short time (ten epochs), and once the termination criteria were met, they trained the final solution for 1000 epochs. Therefore, in line with the parameters used in the referred paper, we set the parameters of our algorithm according to Table~\ref{table:test-ii-params}. 

\begin{table}[h]
    \caption{Parameters of E1.ii: RESN vs. Gradient-based architecture optimization}
	\centering 
    \begin{tabular}{ lrc|clr }
    \hline
    Param & Value &
        && Param & Value \\
                \hline
    cell\_mut\_p & 0.2 &
        && min\_LB & 2 \\
    epochs & 1000 &
        && max\_LB & 30 \\
    max\_step & 15 &
        && min\_NPL & 10 \\
    layer\_mut\_p & 0.2 &
        && max\_NPL & 300 \\
    population\_size & 10 &
        && min\_HL & 1 \\
    offspring\_size & 10 &
        && max\_HL & 8 \\
    max\_eval & 100 &
        && & \\
    \hline
	\end{tabular} \\    
    \label{table:test-ii-params}
\end{table}

And third, as a sanity check, we propose to (\emph{E1.iii}) optimize the architecture (constrained to the architecture search space defined in Table~\ref{table:test-ii-params}) using a \emph{Random Search} algorithm.

Note that in all cases, we propose to split the data sets into training (64\% of the data), validation (16\%), and test (20\%) data. Also, we set an early
stop criteria for the \emph{training} of the final solution (Algorithm~\ref{algorithm:optimizer}, line 14), thus when the validation loss is below 1e-5 the training is stopped, and we propose to add a 0.5 dropout to that training.

\subsubsection{E2: RESN vs. Neuroevolution}

To answer the question of whether a \emph{hybrid} architecture optimization technique can get a similar error performance of a non-training-based approach, we propose to (\emph{E2}) benchmark our approach against EXALT~\cite{elsaid2019evolving}, a state-of-the-art neuroevolutionary technique, on the coal-fired prediction problem. Also, we compare qualitative our results against EXAMM~\cite{ororbia2019investigating} (on the coal-fired prediction problem). Note that we did not propose a direct quantitative comparison because EXAMM evolves the architecture as well as the cell structures. Thus the problem being solved is slightly different. Moreover, in EXAMM the authors used a different experimental design (compared to~\cite{elsaid2019evolving}).

Particularly, we replicate the experimental design of EXALT~\cite{elsaid2019evolving}. 
Specifically in this experiment, we use \emph{K}-fold cross validation, using each burner as a test case (i.e., $K=12$). Additionally, accordingly to EXALT experimental design, we use \emph{stochastic gradient descent} (instead of \emph{Adam}) to train the networks (Algorithm~\ref{algorithm:optimizer}, line 14) with a learning rate $\eta = 0.001$, utilizing Nesterov momentum with \emph{mu}$=0.9$, and without dropout. Also, we use gradient clipping when the norm of the gradient was above 1.0, and boosting when the norm of the gradient was below 0.05. Finally, we set RESN parameters according to Table~\ref{table:rq2-test-params}. Note that the number of epochs of training of the final solution was defined considering the experimental design of EXALT~\cite{elsaid2019evolving}.

\begin{table}[h]
    \caption{Parameters of E2: RESN vs. Neuroevolution}
	\centering 
    \begin{tabular}{ lrc|clr }
    \hline
    Param & Value &
        && Param & Value \\
                \hline
    cell\_mut\_p & 0.2 &
        && min\_LB & 2 \\
    epochs & 1000 &
        && max\_LB & 30 \\
    max\_step & 5 &
        && min\_NPL & 1 \\
    layer\_mut\_p & 0.2 &
        && max\_NPL & 100 \\
    population\_size & 10 &
        && min\_HL & 1 \\
    offspring\_size & 10 &
        && max\_HL & 8 \\
    max\_eval & 100 &
        && & \\
    \hline
	\end{tabular} \\    
    \label{table:rq2-test-params}
\end{table}

\subsubsection{E3: Optimization Time}

As to answer RQ3, we propose to (\emph{E3}) record the execution (run) time of E1 (because we do not have the execution times of E2 competitors), and use it to benchmark our proposal against the training-based architecture optimization techniques.

\subsubsection{E4: RESN vs. Expert Design}

To answer the question of whether RESN can improve the performance of an expert designed architecture/solution (RQ4), we propose to (E4) benchmark our proposal against the winning solution of the ``Electricity Load Forecast using Intelligent Adaptive Technologies''~\cite{chen2004load}, and with state-of-the-art approaches evaluated on this challenge~\cite{lang2018short}.

In this case, we set RESN parameters to their \emph{defaults} (Table~\ref{table:rq4-test-params}). Also, we normalized the data to have a mean equal to zero and a standard deviation equal to one, according to the pre processing suggested by Lang et al.~\cite{lang2018short}, and we set the activation function of the output layer to be \texttt{linear}.

\begin{table}[h]
    \caption{Parameters of E4: RESN vs. Expert design}
	\centering 
    \begin{tabular}{ lrc|clr }
    \hline
    Param & Value &
        && Param & Value \\
                \hline
    cell\_mut\_p & 0.2 &
        && min\_LB & 2 \\
    epochs & 1000 &
        && max\_LB & 30 \\
    max\_step & 5 &
        && min\_NPL & 1 \\
    layer\_mut\_p & 0.2 &
        && max\_NPL & 100 \\
    population\_size & 10 &
        && min\_HL & 1 \\
    offspring\_size & 10 &
        && max\_HL & 3 \\
    max\_eval & 100 &
        && & \\
    \hline
	\end{tabular} \\    
    \label{table:rq4-test-params}
\end{table}

\section{Experimental Results}\label{section:results-optimization}

To carry out the experiments defined in Section~\ref{section:setup}, we have implemented RESN in Python (code available in \url{https://github.com/acamero/dlopt}), using DLOPT~\cite{camero2018dlopt}, Keras~\cite{chollet2015keras}, and Tensorflow~\cite{abadi2016tensorflow}. Then, we run the experiments on the defined problems. Particularly, we repeated each experiment 30 independent times, and we computed the statistics of the error over the final solution.

First, we addressed \emph{E1.i}. Table~\ref{table:results-rq1-i} summarizes the results of the \emph{E1.i} experiment, where RESN stands for the optimization guided by the MRS heuristic and GDET for Adam training heuristc (defined in Section~\ref{section:setup}). The best mean and median are in bold font.

\begin{table}[!h]
    \caption{E1.i results on the sine wave problem (MAE of the final solution)}
	\centering 
    \begin{tabular}{ lrr }
    \hline
       & RESN & GDET    \\
    \hline
    Mean    & \textbf{0.105}     & 0.142   \\
    Median  & \textbf{0.100}     & 0.149   \\
    Max     & 0.247     & 0.270   \\
    Min     & 0.063     & 0.054   \\
    Sd      & 0.035     & 0.051   \\
    \hline
	\end{tabular} \\    
    \label{table:results-rq1-i}
\end{table}

\begin{table*}[!t]
    \caption{E1.ii results. RNN optimization in the Waste generation prediction problem, a comparison between Short training and RESN}
	\centering
    \begin{tabular}{ lrrrrrr|rrrrrr }
    \toprule
       & \multicolumn{5}{c}{\textit{Short training}} &&& \multicolumn{5}{c}{\textit{RESN}}  \\ 
       \cline{2-6} \cline{9-13} 
       & MAE & No. LSTM & LB & No. HL & Time [min]
       &&& MAE & No. LSTM & LB & No. HL & Time [min] \\ 
    \hline
    Mean    & \textbf{0.073} & 451 & 6 & 5 & 97      &&& 0.079 & 793 & 17 & 3 & \textbf{51} \\
    Median  & \textbf{0.073} & 420 & 5 & 5 & 70      &&& \textbf{0.073} & 513 & 16 & 2 & \textbf{45} \\
    Max     & 0.076 & 1252 & 16 & 8 & 405   &&& 0.138 & 2038 & 30 & 8 & 103 \\
    Min     & 0.071 & 127 & 2 & 1 & 33      &&& 0.069 & 444 & 2 & 1 & 40 \\
    Sd      & 0.001 & 228 & 2 & 2 & 75      &&& 0.017 & 493 & 11 & 3 & 13 \\
    \bottomrule
	\end{tabular} \\    
    \label{table:results-rq1-ii}
\end{table*}

Overall, the results of RESN exceed GDET, i.e., the optimized RNN obtained by RESN (guided by the MRS) has on average a lower error than the ones optimized by GDET. Moreover, the Wilcoxon rank-sum test $p$-value is 0.001. Therefore, we can conclude that RESN is significantly better than GDET.

To continue with the validation of RESN, we executed the \emph{E1.ii} experiment. Table~\ref{table:results-rq1-ii} summarizes the results presented in~\cite{Camero2018Waste} (columns under \emph{Short training}), and our results (columns under RESN). In the table, \emph{MAE} stands for the MAE of the final solution, \emph{No. LSTM} is the number of LSTM in the network, \emph{LB} corresponds to the look back, \emph{No. HL} represents the number of hidden layers, and \emph{Time} is the total time (i.e., the optimization process and the training of the final solution) in minutes. The best mean and median are in bold font.


In terms of the MAE, the results of both approaches (RESN and \emph{short training}) are similar. Therefore, we performed a Wilcoxon rank-sum test to validate if there is a significant difference between them. Note that both approaches are stochastic and were executed 30 independent times for statistical soundness. The $p$-value of the test (comparing the MAE) is equal to 0.665, therefore there is no evidence that one algorithm \emph{outperforms} the other. Furthermore, the median is the same in both cases.

Again, it is important to notice that Ferrer and Alba~\cite{Ferrer2018} originally proposed to predict the filling level of each container individually using \emph{Gaussian processes}, \emph{linear regression}, and \emph{SMReg}. But later, \emph{Short training}~\cite{Camero2018Waste} outperformed those results. Then, our proposal beats all the techniques used in \cite{Ferrer2018}.

On the other hand, RESN (by using the MRS as the heuristic for optimizing the network) dramatically reduces the time needed to optimize the RNN configuration (RQ3). On average, the time has been cut in half (nearly one hour difference). Again, notice that Table~\ref{table:results-rq1-ii} presents the time in minutes.

We run the \emph{E1.iii} experiment. On average, the MAE of the solution found using the random search was equal to 0.091 for the waste prediction problem, and the training time of each network was 25 minutes. Therefore, evaluating 100 networks (\emph{max\_eval}) took nearly 50x (on average) the time needed to optimize the network using RESN.
We also performed a Wilcoxon rank-sum test to compare the Random search against RESN and Short training. The $p$-values are 0.017 and 0.002 respectively. Therefore, we concluded that the Random search does not perform as well as the other techniques. Nonetheless, it is quite interesting that a simple random search could give such good results (MAE). However, its main drawback is the time needed to get a \emph{competitive} solution. 

Later, we run the E2 experiment. Table~\ref{table:results-rq2} presents the mean square error (MSE) of the solution obtained by RESN in the coal-fire power plan problem, as well as the results presented in EXALT~\cite{elsaid2019evolving}. The results show that RESN improves the results of EXALT, with a difference of one order of magnitude. The best result for each fold and for the average is in bold font. In this case, the Wilcoxon rank-sum test $p$-value is 2.958e-06. Despite the small decimal figures involved, our technique has shown to be ten times more accurate than the state-of-the-art, and the statistical tests confirm that this difference is meaningful.

\begin{table}[!t]
    \caption{E2 results (MSE). RESN vs EXALT in the coal-fire power plant problem}
	\centering
    \begin{tabular}{ lrr }
    \hline
    Fold & EXALT & RESN \\
    \hline
    0 & 0.028749 & \textbf{0.001541} \\
    1 & 0.031769 & \textbf{0.006536} \\
    2 & 0.023095 & \textbf{0.003821} \\
    3 & 0.019229 & \textbf{0.000570} \\
    4 & 0.023170 & \textbf{0.003336} \\
    5 & 0.036091 & \textbf{0.000617} \\
    6 & \textbf{0.012879} & 0.017061 \\
    7 & 0.019358 & \textbf{0.004032} \\
    8 & 0.018151 & \textbf{0.001912} \\
    9 & 0.019475 & \textbf{0.013996} \\
    10 & 0.030016 & \textbf{0.006120} \\
    11 & 0.031207 & \textbf{0.002942} \\
    \hline
    Average & 0.024432 & \textbf{0.005208} \\
    \bottomrule
	\end{tabular} \\    
    \label{table:results-rq2}
\end{table}

Then, we processed the results of EXAMM~\cite{ororbia2019investigating}. Particularly, we computed the average MSE for RNNs Evolved With Individual Memory Cells and RNNs Evolved With All Memory Type (Table 1 in the cited article) and for RNNs Evolved With Simple Neurons and Memory Cells (Table 2 in the cited article). The values are 0.001690 and 0.001601, respectively. Although the experiments are not comparable (i.e., EXAMM uses a different experimental design, but the same data set), it is quite interesting to notice that RESN achieves a result (0.005208) that is in the same order of magnitude that EXAMM, but with a much simpler approach, i.e., we only use fully connected stacked LSTM layers, instead of multiple types of cells.

\begin{table*}[!t]
    \caption{E4 results (MAPE). RESN vs Expert design}
	\centering
    \begin{tabular}{ lrrrrrrrr }
    \hline
            & 	SVM     & 	BP  & 	RBF & 	SVR & NNRW  & KNNRW & WKNNRW & 	RESN \\
    \hline
    Mean    & 	2.879   & NA    & NA    & NA    & NA    & NA    & NA    & 	2.281 \\ 
    Median  & 	2.945   & NA    & NA    & NA    & NA    & NA    & NA    & 	2.242 \\ 
    Max     & 	3.480   & NA    & NA    & NA    & NA    & NA    & NA    & 	3.273 \\ 
    Min     & 	1.950   & 1.451 & 1.481 & 1.446 & 1.438 & 1.348 & 1.323 & 	1.370 \\ 
    Std     & 	0.004   & NA    & NA    & NA    & NA    & NA    & NA    & 	0.414 \\ 
    \bottomrule
	\end{tabular} \\    
    \label{table:results-rq4}
\end{table*}

Finally, we run the E4 experiment. In this case, we reported the mean absolute percentage error (MAPE), as it is the metric reported in the benchmarked studies.The summarized results are presented in Table~\ref{table:results-rq4}. The column SVM corresponds to the results presented by Chen et al.~\cite{chen2004load}, the winner of the ``Electricity Load Forecast using Intelligent Adaptive Technologies'' competition organized by EUNITE. In the referred study, the authors proposed a support vector machine approach to predict the load. It is important to notice that they also studied in deep the data set, and proposed several preprocessings to the prepare the data for the specific task.

On the other hand, the columns BP, RBF, SVR, NNRW, KNNRW, and WKNNRW correspond to the results presented by Lang et al.~\cite{lang2018short}. In this study, the authors proposed several approaches to tackle the load forecast problem. Particularly, they proposed to use BP (backpropagation neural network), RBF (radial basis function network), SVR (support vector regression), NNRW (neural network with random weights), KNNRW (neural network with random weights and kernels), and WKNNRW (weighted neural network with random weights and kernels). As a remark, the authors did not report statistical results, thus we added a NA (not available) to the missing data.

Overall, the performance of RESN is comparable to a human expert, meaning that the error of the best solution found is as good as the best solution proposed by the experts. We acknowledge that the techniques are not the same, thus there may be a bias in this benchmark. Nonetheless, the comparison is \emph{fair} in terms that all the considered approaches were specifically tailored for the load forecast problem, and that all of them were tuned manually with expert domain/technique knowledge.

As a summary, RESN has a similar (or better) error performance to training-based RNN optimization techniques (RQ1) and to neuroevolutionary approaches (RQ2) but considerably reduces the computational time (RQ3). Moreover, the performance of RESN is as good as the best solution proposed by an expert (RQ4). Hence, we offer a competitive alternative to architecture optimization that does not rely on training but on the MRS, and that does not require expert domain/technique knowledge to tailor a solution a specific problem.

\section{Conclusions and Future Work}\label{section:conclusions}

In this work, we present RESN, an EA to optimize the architecture of an RNN that uses the MRS as a search heuristic. We have evaluated our proposal on four prediction problems and compared our results against training-based architecture optimization techniques, neuroevolutionary approaches, and human expert designed approaches.

The results show that RESN is as good as a training-based architecture optimization technique in terms of the error, i.e., the prediction error of our solutions are similar to the ones of the architectures optimized using training-based architecture optimization techniques. 

When comparing  RESN against \emph{pure} neuroevolutionary techniques, the results show that RESN achieves state-of-the-art performance.  
Moreover, the evidence presented in this work shows that our approach reduces by half the time needed to optimize the architecture of an RNN compared to a training-based architecture optimization technique (on average, compared to short training results, we reduce the time from 97 to 51 minutes in the waste generation prediction problem, and RESN is 50x faster than training each candidate architecture). Therefore, in the future, we could optimize much larger networks in the time that existing algorithms need for smaller cases.

Moreover, the results show that RESN is a competitive replacement to human expert design solutions, with the add-on that no domain specific knowledge is required to tailor the RNN to the problem.

According to these results, we conclude that RESN provides a competitive alternative for RNN optimization. Overall, the results suggest that the MRS is a promising method to compare RNN architectures and that it is a very useful heuristic for architecture optimization.

As future work, we propose to extend our proposal (and the MRS) to other problem classes, e.g., classification, clustering, among others. Moreover, we will analyze the use of other metaheuristics, such as, a GA with specific operators, that will allow to improve the search of network architectures. On the other hand, we envision the importance of studying thoroughly the configuration of the MRS parameters, including using a different PDF, training algorithm, among others.

\subsubsection*{Acknowledgments}
This research was partially funded by Universidad de M\'alaga, Andaluc\'ia Tech, Consejer\'ia de Econom\'ia y Conocimiento de la Junta de Andalu\'ia, Ministerio de Econom\'ia, Industria y Competitividad, Gobierno de Espa\~na, and European Regional Development Fund grant numbers TIN2017-88213-R (6city.lcc.uma.es), RTC-2017-6714-5 (ecoiot.lcc.uma.es), and UMA18-FEDERJA-003 (Precog). 
European Union’s Horizon 2020 research and innovation programme under the Marie Skłodowska-Curie grant agreement No 799078. 
Helmholtz Association’s Initiative and Networking Fund (INF) under the Helmholtz AI platform grant agreement (ID ZT-I-PF-5-1).


\balance

\bibliographystyle{elsarticle-num-names}

\bibliography{bibliography}

\end{document}